\documentclass[10pt,twocolumn,letterpaper]{article}
\usepackage{geometry}
\usepackage{cvpr}
\usepackage{times}
\usepackage{epsfig}
\usepackage{graphicx}
\usepackage{amsmath}
\usepackage{amssymb}

\usepackage[linesnumbered, ruled]{algorithm2e}
\usepackage{algorithmic}
\usepackage[utf8]{inputenc} 
\usepackage[T1]{fontenc}    
\usepackage{url}            
\usepackage{booktabs}       
\usepackage{amsfonts}       
\usepackage{nicefrac}       
\usepackage{microtype}      
\usepackage{amsmath,amssymb}
\usepackage{color}
\usepackage{float}
\usepackage{array}
\usepackage{multirow}
\usepackage{booktabs}
\usepackage{subfigure}
\usepackage{amsthm}
\usepackage{wrapfig,lipsum,booktabs}
\usepackage{xcolor}
\usepackage{bm}
\usepackage{textcomp}
\usepackage[noblocks]{authblk}
\newcommand{\red}[1]{{\color{red}#1}}
\newcommand{\blue}[1]{{\color{blue}#1}}

\newcommand{\Koven}{\color{black}}
\newcommand{\jason}[1]{{\color{black}#1}}

\newcommand{\erhao}{\fontsize{21pt}{\baselineskip}\selectfont}
\newtheorem{assump}{Assumption}

\usepackage[breaklinks=true,bookmarks=false]{hyperref}

\cvprfinalcopy 

\def\cvprPaperID{522} 

\ifcvprfinal\fi
\begin{document}

{\onecolumn

\noindent \textbf{\erhao{Unsupervised Person Re-identification by Soft Multilabel Learning}}

\vspace{2cm}

\noindent {\LARGE{Hong-Xing Yu, Wei-Shi Zheng, \\Ancong Wu, Xiaowei Guo, Shaogang Gong, Jian-Huang Lai}}

\vspace{2cm}

\noindent Code is available at: \\
\ \ \ \ \ \ \ \ \ \ \ \ \url{https://github.com/KovenYu/MAR}

\vspace{1cm}

\noindent For reference of this work, please cite:

\vspace{1cm}
\noindent Hong-Xing Yu, Wei-Shi Zheng, Ancong Wu, Xiaowei Guo, Shaogang Gong and Jian-Huang Lai.
``Unsupervised Person Re-identification by Soft Multilabel Learning''
In \emph{Proceedings of the IEEE International Conference on Computer Vision and Pattern Recognition (CVPR).} 2019.

\vspace{1cm}

\noindent Bib:\\
\noindent
@inproceedings\{yu2019unsupervised,\\
\ \ \   title=\{Unsupervised Person Re-identification by Soft Multilabel Learning\},\\
\ \ \  author=\{Yu, Hong-Xing and Zheng, Wei-Shi and Wu, Ancong and Guo, Xiaowei and Gong, Shaogang and Lai, Jian-Huang\},\\
\ \ \  booktitle=\{Proceedings of the IEEE International Conference on Computer Vision and Pattern Recognition (CVPR)\},\\
\ \ \  year=\{2019\}\\
\}
}

%
\restoregeometry

\title{Unsupervised Person Re-identification
by {\Koven Soft Multilabel Learning}}


\author[ ]{\vspace{-0.9cm}Hong-Xing Yu$^1$}
\author[ ]{Wei-Shi Zheng$^{1,4}$\thanks{Corresponding author}}
\author[ ]{\\Ancong Wu$^1$}
\author[ ]{Xiaowei Guo$^2$}
\author[ ]{Shaogang Gong$^3$}
\author[ ]{Jian-Huang Lai$^1$\vspace{0cm}}

\affil[ ]{$^1$Sun Yat-sen University, China}
\affil[ ]{$^2$YouTu Lab, Tencent$\quad\quad^3$Queen Mary University of London, UK}
\affil[ ]{$^4$Key Laboratory of Machine Intelligence and Advanced Computing, Ministry of Education, China}
\affil[ ]{\tt\small xKoven@gmail.com, wszheng@ieee.org, wuancong@mail2.sysu.edu.cn, scorpioguo@tencent.com, s.gong@qmul.ac.uk, stsljh@mail.sysu.edu.cn}

\maketitle
\thispagestyle{empty}

\begin{abstract}
Although unsupervised person re-identification (RE-ID) has drawn increasing research attentions
due to its potential to address the scalability problem of supervised RE-ID models,
it is very challenging to learn discriminative information in the absence of pairwise labels across disjoint camera views.
To overcome this problem, we propose a deep model for the soft multilabel learning for unsupervised RE-ID.
The idea is to learn a soft multilabel
(real-valued label likelihood vector) for each unlabeled person
by comparing (and representing) the unlabeled person with a set of known \emph{reference} persons from an auxiliary domain.
We propose the soft multilabel-guided hard negative mining to learn a discriminative embedding for
the unlabeled target domain by exploring the \emph{similarity consistency} of the visual features and the soft multilabels
of unlabeled target pairs.
Since most target pairs are cross-view pairs, we develop the cross-view consistent soft multilabel learning
to achieve the learning goal that the soft multilabels are consistently good across different camera views.
To enable effecient soft multilabel learning, we introduce the reference agent learning
to represent each reference person by a reference agent in a joint embedding.
We evaluate our unified deep model on Market-1501 and DukeMTMC-reID.
Our model outperforms the state-of-the-art unsupervised RE-ID
methods by clear margins.
Code is available at \url{https://github.com/KovenYu/MAR}.
\end{abstract}
\vspace{-0.5cm}

\section{Introduction}

Existing person re-identification (RE-ID) works mostly focus on supervised learning
\cite{2012_CVPR_KISSME,2015_CVPR_LOMO,2017_TPAMI_SALIENCE,2018_CVPR_camera-style,2015_CVPR_Ahmed,2016_CVPR_Wang,2016_CVPR_JSTL,
2016_ECCV_gated,2018_CVPR_mask-guided,2018_ECCV_PCB,2018_CVPR_pose-sensitive}.
However, they need substantial pairwise labeled data across every pair of camera views,
limiting the scalability to large-scale applications where only unlabeled data is available due to the prohibitive manual efforts
in exhaustively labeling the pairwise RE-ID data \cite{2015_TCSVT_xiaojuan}.
To address the scalability problem,
some recent works focus on unsupervised RE-ID
by clustering on the target unlabelled data \cite{2017_ICCV_asymmetric,2019_TPAMI_DECAMEL,2017_Arxiv_PUL} or transfering the knowledge from other labeled source dataset
\cite{2016_CVPR_tDIC,2018_CVPR_transferable,2018_CVPR_SPGAN,2018_ECCV_HHL}.
However, the performance is still not satisfactory.
The main reason is that, without the pairwise label as learning guidence,
it is very challenging to discover the identity discriminative information
due to the drastic cross-view intra-person appearance variation \cite{2017_ICCV_asymmetric}
and the high inter-person appearance similarity \cite{2013_TPAMI_RDC}.

\begin{figure}[t]
\begin{center}
\includegraphics[width=0.9\linewidth]{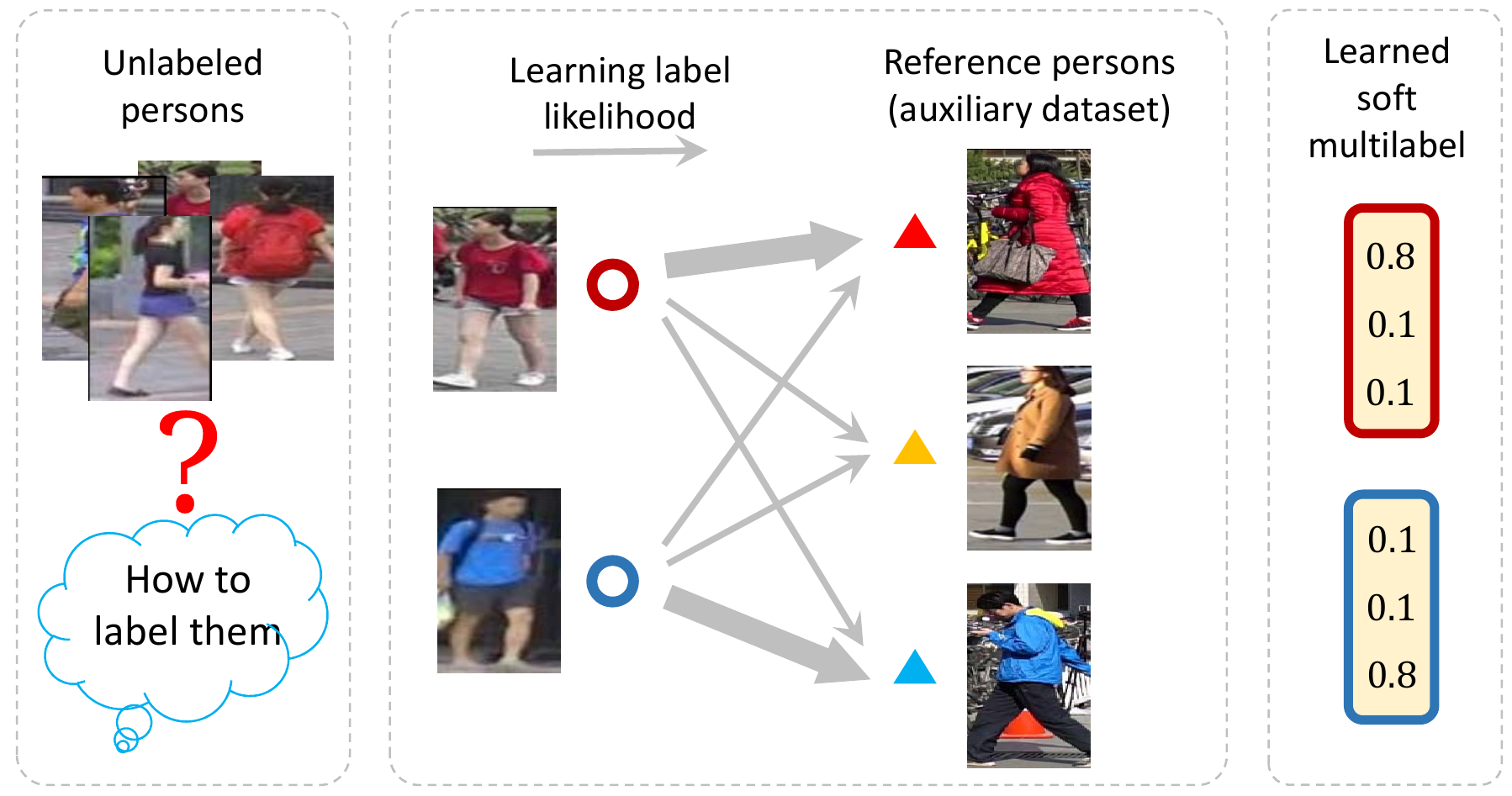}
\end{center}
\vspace{-0.2cm}
   \caption{Illustration of our soft multilabel concept.
   We learn a soft multilabel (real-valued label vector) for each unlabeled person
   by comparing to a set of known auxiliary reference persons
   (thicker arrowline indicates higher label likelihood).
   Best viewed in color.}
\vspace{-0.4cm}
\label{fig:illustration}
\end{figure}

To address the problem of lacking pairwise label guidance in unsupervised RE-ID,
in this work we propose a novel soft multilabel learning
to mine the potential label information in the unlabeled RE-ID data.
The main idea is,
for every unlabeled person image in an unlabeled RE-ID dataset,
we learn a soft multilabel (i.e. a real-valued
label likelihood vector instead of a single pseudo label) by
comparing this unlabeled person with a set of \emph{reference persons} from an existing labeled auxiliary source dataset.
Figure \ref{fig:illustration} illustrates this soft multilabel concept.

Based on this soft multilabel learning concept,
we propose to mine the potential discriminative information by the \emph{soft multilabel-guided hard negative mining},
i.e. we leverage the soft multilabel to distinguish the visually similar but different unlabeled persons.
In essence, the soft multilabel represents the unlabelled target person by
the reference persons,
and thus it encodes the {\em relative} comparative
characteristic of the unlabeled person,
which is from a different perspective than the {\em absolute} visual feature representation.
Intuitively, a pair of images of the same person should be not only visually similar to each other (i.e. they should have similar absolute visual features),
but also equally similar to any other reference person (i.e. they should also have similar relative comparative characteristics with respect to the reference persons).
If this \emph{similarity consistency} between the absolute visual representation and the relative soft multilabel representation is violated,
i.e. the pair of images are visually similar but their comparative characteristics are dissimilar,
it is probably a hard negative pair.

In the RE-ID context, most image pairs are \emph{cross-view} pairs which consist of two person images captured by different camera views.
Therefore, we propose to learn the soft multilabels that are consistently good across different camera views.
We refer to this learning as the \emph{cross-view consistent soft multilabel learning}.
To enable the efficient soft multilabel learning which requires comparison between the unlabeled persons and the reference persons,
we {\Koven introduce} the \emph{reference agent learning} to represent each reference person
by a reference agent which resides in a joint feature embedding with the unlabeled persons.
Specifically, we {\Koven develop} a unified deep model named deep soft \textbf{m}ultil\textbf{a}bel \textbf{r}eference learning (MAR)
which jointly formulates the soft multilabel-guided hard negative mining, the cross-view consistent soft multilabel learning
and the reference agent learning.

We summarize our \textbf{contributions} as follows:
(1). We address the unsupervised RE-ID problem
by a novel soft multilabel reference learning method,
in which we mine the potential label information {\em latent} in the unlabeled RE-ID data
by exploiting the auxiliary source dataset for reference comparison.
(2). We formulate a novel deep model named deep soft
\textbf{m}ultil\textbf{a}bel \textbf{r}eference learning (MAR).
MAR enables simultaneously the soft multilabel-guided hard negative mining,
the cross-view consistent soft multilabel learning and the reference agent learning in a unified model.
Experimental results on Market-1501 and DukeMTMC-reID show that our model outperforms the
state-of-the-art unsupervised RE-ID methods by significant margins.

\section{Related Work}

\noindent
\textbf{Unsupervised RE-ID}.
Unsupervised RE-ID refers to that the target dataset is unlabelled
but the auxiliary source dataset is not necessarily unlabelled
\cite{2016_CVPR_tDIC,2017_Arxiv_PUL,2018_CVPR_transferable,2018_CVPR_SPGAN,2018_ECCV_HHL}.
Existing methods either transfer source label knowledge \cite{2016_CVPR_tDIC,2017_Arxiv_PUL,2018_CVPR_transferable,2018_CVPR_SPGAN,2018_ECCV_HHL}
or assuming strong prior knowledge
(i.e. either assuming the target RE-ID data has specific cluster structure \cite{2017_ICCV_asymmetric,2019_TPAMI_DECAMEL,2017_Arxiv_PUL}
or assuming the hand-crafted features could be discriminative enough \cite{2017_ICCV_asymmetric,Dic,GL,2013_CVPR_saliency,SDALF,GTS}).
Recently attempts have been made on exploiting video tracklet associations for unsupervised RE-ID \cite{2018_BMVC_tracklet,2018_ECCV_tracklet}.
Another line of work focusing on reducing the labelling effort is to minimize the labelling budget on the target \cite{2018_CVPR_transitivity}
which is complementary to the unsupervised RE-ID.
The most related works are the clustering-based models
\cite{2017_ICCV_asymmetric,2019_TPAMI_DECAMEL,2017_Arxiv_PUL},
e.g.
Yu et.al. \cite{2017_ICCV_asymmetric,2019_TPAMI_DECAMEL} proposed an asymmetric metric clustering to discover labels latent
in the unlabelled target RE-ID data.
The main difference
is that the soft multilabel could leverage the auxiliary reference information other than visual feature similarity,
while the pseudo label only encodes the feature similarity of an unlabelled pair.
Hence, the soft multilabel could mine the potential label information
that cannot be discovered by directly comparing the visual features.

Some unsupervised RE-ID works also proposed to use the labeled source dataset
by the unsupervised domain adaptation \cite{2018_CVPR_PTGAN,2018_CVPR_SPGAN,2018_ECCV_HHL,2018_CVPR_transferable}
to transfer the discriminative knowledge from the auxiliary source domain.
Our model is different from them in that
these models do not mine the discriminative information in the unlabeled target domain,
which is very important because the transferred discriminative knowledge might be less effective
in the target domain due to the domain shift \cite{2010_TKDE_survey} in discriminative visual clues.

\vspace{0.1cm}
\noindent
\textbf{Unsupervised domain adaptation}.
Our work is also closely related to the unsupervised domain adaptation (UDA)
\cite{2015_ICML_DAN,2016_AAAI_CORAL,2016_ECCVW_CORAL,2018_ICLR_DIRTY,2018_ICLR_EM,2017_CVPR_ADDA,2015_ICML_backpropagation,2017_ICML_JAN},
which also has a source dataset
and an unlabeled target dataset.
However, they are mostly based on the assumption that the classes are the same between both domains \cite{2010_ML_bendavid,2010_TKDE_survey,2017_Arxiv_DA-survey,2011_TNN_TCA},
which does not hold in the RE-ID context
where the persons (classes) in the source dataset are completely different from the persons in the target dataset,
rendering these UDA models inapplicable to the unsupervised RE-ID \cite{2018_CVPR_PTGAN,2018_CVPR_SPGAN,2018_ECCV_HHL,2018_CVPR_transferable}.

\vspace{0.1cm}
\noindent
\textbf{Multilabel classification}.
Our soft multilabel learning is conceptually different from the multilabel classification \cite{2014_TKDE_multilabel}.
The multilabel in the multilabel classification \cite{2014_TKDE_multilabel} is a groundtruth binary vector indicating
whether an instance \emph{belongs to} a set of classes,
while our soft multilabel is learned to represent an unlabeled target person
by \emph{other different} reference persons.
{\Koven
Hence, existing multilabel classification models are for a different purpose and thus not suitable to model our idea.
}

\vspace{0.1cm}
\noindent
\textbf{Zero-shot learning}.
Zero-shot learning (ZSL) aims to recognize novel testing classes specified by semantic attributes but unseen during training \cite{2009_CVPR_ZSL,2015_ICML_ZSL,2015_ICCV_ZSL,2015_ICCV_UDA4ZSL,2016_CVPR_ZSL}.
Our soft multilabel reference learning is related to ZSL in that
every unknown target person (unseen testing class) is represented by a set of known
reference persons (attributes of training classes).
However, the predefined semantic attributes are not available in unsupervised RE-ID.
Nevertheless, the success of ZSL models validates/justifies the effectiveness of
representing an unknown class (person) with a set of different classes.
A recent work also explores a similar idea by representing an unknown testing person in an ID regression space
which is formed by the known training persons \cite{2018_IJCV_ID-regression},
but it requires substantial labeled persons from the target domain.

%

\section{Deep Soft Multilabel Reference Learning}

\subsection{Problem formulation and Overview}

\begin{figure*}[t]
\begin{center}
\includegraphics[width=0.8\linewidth]{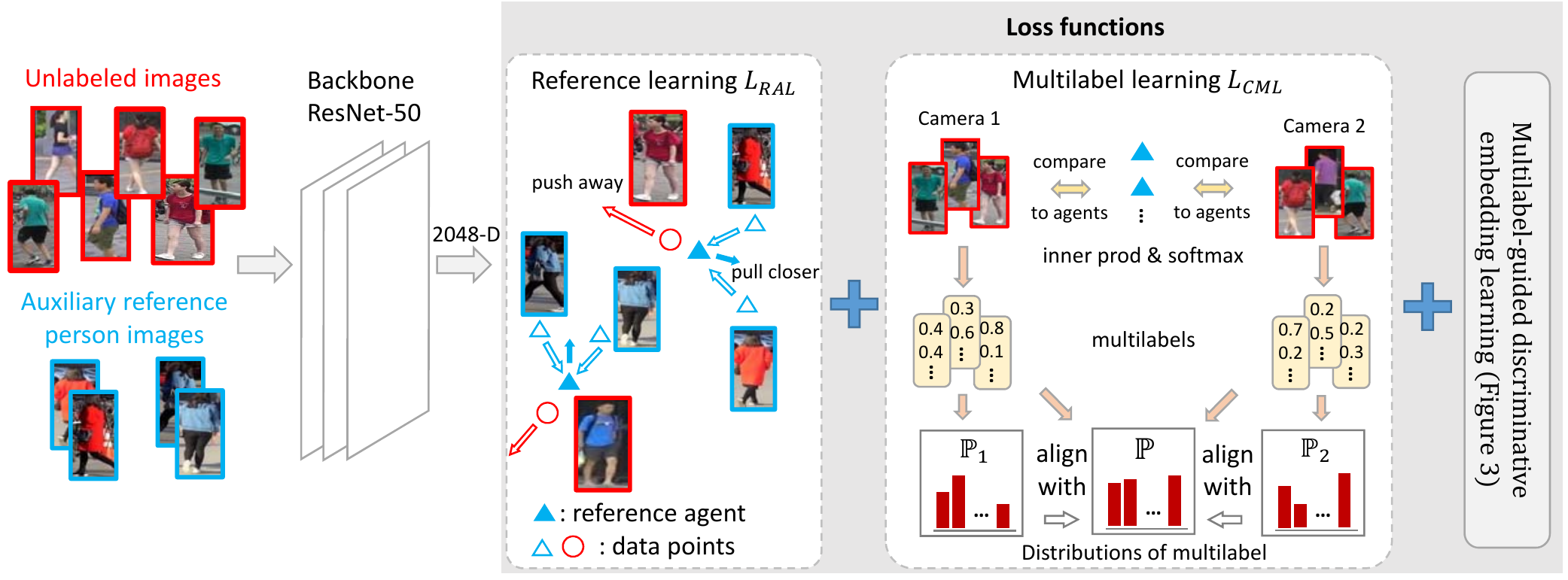}
\end{center}
\vspace{-0.2cm}
   \caption{An illustration of our model MAR.
   We learn the soft multilabel by comparing each target unlabeled person
   image $f(x)$ (red circle) to a set of auxiliary reference persons
   represented by a set of reference agents $\{a_i\}$ (blue triangles, learnable parameters) in the feature embedding.
   The soft multilabel judges whether a similar pair is positive or hard negative for discriminative embedding learning (Sec. \ref{sec:hard_negative}).
   The soft multilabel learning and the reference learning are elaborated in Sec. \ref{sec:soft multilabel_learning} and Sec. \ref{sec:aj}, respectively.
   Best viewed in color.}
\vspace{-0.2cm}
\label{fig:framework}
\end{figure*}

We have an unlabeled target RE-ID dataset $\mathcal{X} =
\{x_i\}_{i=1}^{N_u}$ where each $x_i$ is an unlabeled person image
collected in the target visual surveillance scenario,
and an auxiliary RE-ID dataset $\mathcal{Z} = \{z_i,
w_i\}_{i=1}^{N_a}$ where each $z_i$ is a person image
with its label $w_i = 1, \cdots, N_p$ where $N_p$ is the number of the
reference persons.
Note that the reference population is completely
non-overlapping with the unlabeled target population since it is
collected from a different surveillance scenario
\cite{2018_CVPR_PTGAN,2018_CVPR_SPGAN,2018_ECCV_HHL}.
Our goal is to learn a soft multilabel function $l(\cdot)$ such that $y = l(x, \mathcal{Z}) \in (0,1)^{N_p}$ where
all dimensions add up to 1 and each dimension represents the label likelihood w.r.t. a reference person.
Simultaneously, we aim to learn a discriminative deep feature embdding $f(\cdot)$ under the guidance of the soft multilabels for the RE-ID task.
Specifically, we propose to leverage the soft multilabel for hard negative mining,
i.e. for visually similar pairs we determine they are positive or hard negative
by comparing their soft multilabels.
We refer to this part as the \emph{Soft multilabel-guided hard negative mining} (Sec. \ref{sec:hard_negative}).
In the RE-ID context, most pairs are \emph{cross-view} pairs which consist of two person images captured by different camera views.
Therefore, we aim to learn the soft multilabels that are consistently
good across different camera views
so that the soft multilabels of the cross-view images are comparable.
We refer to this part as the \emph{Cross-view consistent soft multilabel learning} (Sec. \ref{sec:soft multilabel_learning}).
To effeciently compare each unlabeled person $x$ to all the reference persons,
we introduce the \emph{reference agent learning} (Sec. \ref{sec:aj}),
i.e. we learn a set of \emph{reference agents} $\{a_i\}_{i=1}^{N_p}$
each of which represents a reference person in the shared \emph{joint feature embedding}
where both the unlabeled person $f(x)$ and the agents $\{a_i\}_{i=1}^{N_p}$ reside (so that they are comparable).
Therefore, we could learn the soft multilabel $y$ for $x$ by comparing $f(x)$ with the reference agents $\{a_i\}_{i=1}^{N_p}$,
i.e. the soft multilabel function is simplified to $y = l(f(x), \{a_i\}_{i=1}^{N_p})$.

We show an overall illustration of our model in Figure \ref{fig:framework}.
In the following, we introduce our deep soft \textbf{m}ultil\textbf{a}bel \textbf{r}eference learning (MAR).
We first introduce the soft multilabel-guided hard negative mining
given the reference agents $\{a_i\}_{i=1}^{N_p}$ and the reference comparability between $f(x)$ and $\{a_i\}_{i=1}^{N_p}$.
To facilitate learning the joint embedding, we enforce a unit norm constraint,
i.e. $||f(\cdot)||_2=1, ||a_i||_2=1, \forall i$, to learn a hypersphere embedding \cite{2017_ACMMM_normface,2017_CVPR_sphereface}.
Note that in the hypersphere embedding, the cosine similarity between a pair of features $f(x_i)$ and $f(x_j)$
is simplified to their inner product $f(x_i)^\mathrm{T}f(x_j)$,
and so as for the reference agents.

\subsection{Soft multilabel-guided hard negative mining}\label{sec:hard_negative}
{\Koven Let us start by defining the soft multilabel function.}
Since each entry/dimension of the soft multilabel $y$ represents the label likelihood that adds up to 1,
we define our soft multilabel function as
{\small
\begin{align}\label{eq:ML}
y^{(k)} = l(f(x), \{a_i\}_{i=1}^{N_p})^{(k)} = \frac{\exp(a_k^\mathrm{T}f(x))}{\Sigma_{i} \exp(a_i^\mathrm{T}f(x))}
\end{align}
}%
where $y^{(k)}$ is the $k$-th entry of $y$.

It has been shown extensively that mining hard negatives
is more important in learning a discriminative embedding than naively learning from all
visual samples \cite{2017_Arxiv_triplet-loss,2016_NIPS_tuplet-loss,2016_CVPR_lifted-structured,2016_ECCV_moderate-mining,2015_CVPR_facenet}.
We explore a soft multilabel-guided hard negative mining,
which focuses on the pairs of visually similar but different persons and aims to distinguish them with the guidance of their soft multilabels.
Given that the soft multilabel encodes relative comparative characteristics,
we explore a representation consistency:
Besides the similar \emph{absolute} visual features,
images of the same person should also have similar \emph{relative} comparative characteristics
(i.e. equally similar to any other reference person).
Specifically, we make the following assumption in our model formulation:

\begin{assump}\label{assump}
If a pair of unlabeled person images $x_i, x_j$ has high feature similarity $f(x_i)^\mathrm{T}f(x_j)$,
we call the pair a \emph{similar pair}.
If a similar pair has highly similar comparative characteristics, it is probably a positive pair.
Otherwise, it is probably a hard negative pair.
\end{assump}

For the similarity measure of the comparative characteristics encoded in the pair of soft multilabels,
we propose the \emph{soft multilabel agreement} $A(\cdot, \cdot)$,
defined by:
{\footnotesize
\begin{align}\label{eq:agreement}
A(y_i, y_j) = y_i \wedge y_j = \Sigma_k \min(y_i^{(k)}, y_j^{(k)}) =  1-\frac{||y_i-y_j||_1}{2},
\end{align}
}%
 which is based on the well-defined L1 distance.
Intuitively, the soft multilabel agreement is an analog to the voting by the reference persons:
Every reference person $k$ gives his/her conservative agreement
$\min(y_i^{(k)}, y_j^{(k)})$ on believing the pair to be positive
(the more similar/related a reference person is to the unlabeled pair, the more important is his/her word),
and the soft multilabel agreement is cumulated from all the reference persons.
{\Koven The soft multilabel agreement is defined based on L1 distance to treat fairly the agreement of every reference person by taking the absolute value.
}

\begin{figure}[t]
\begin{center}
\includegraphics[width=1\linewidth]{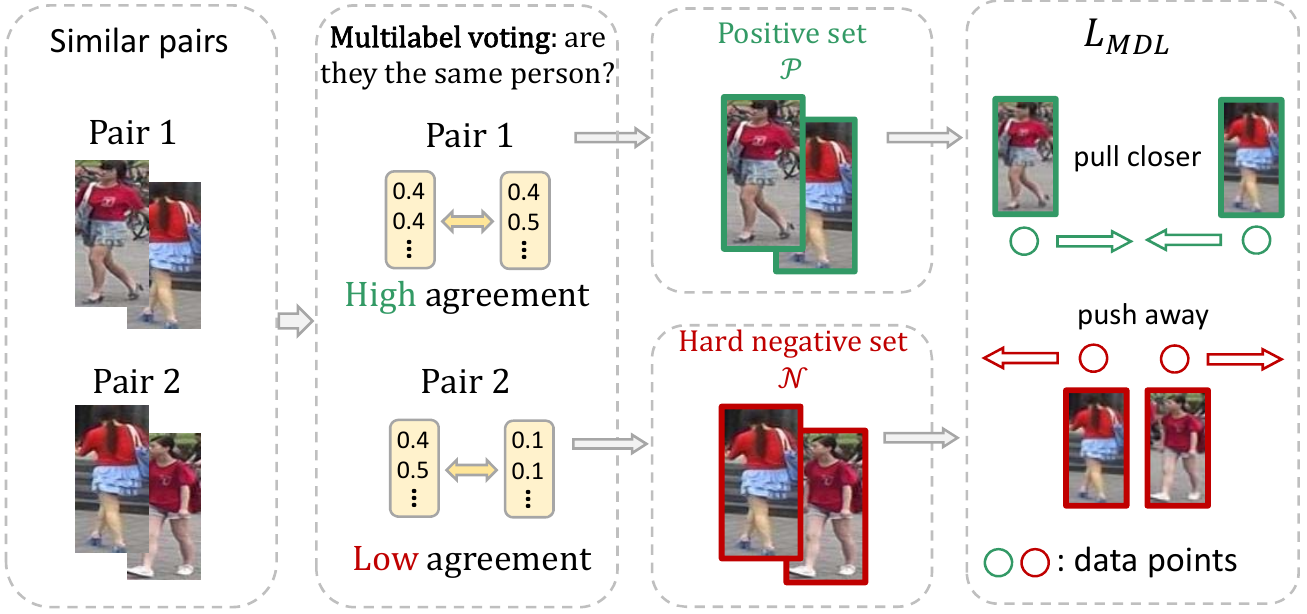}
\end{center}
\vspace{-0.2cm}
   \caption{Illustration of the soft multilabel-guided hard negative mining.
   Best viewed in color.
   }
\vspace{-0.2cm}
\label{fig:MDL}
\end{figure}

Now, we mine the hard negative pairs by considering both the feature
similarity and soft multilabel agreement according to Assumption \ref{assump}.
We formulate the soft multilabel-guided hard negative mining with a mining ratio $p$:
We define the similar pairs in Assumption \ref{assump} as the $pM$ pairs
that have highest feature similarities among all the
$M=N_u\times(N_u-1)/2$ pairs within the unlabeled target dataset $\mathcal{X}$.
For a similar pair $(x_i,x_j)$, if it is also among the top $pM$ pairs that have the highest soft multilabel agreements,
we assign $(i,j)$ to the positive set $\mathcal{P}$, otherwise we
assign it to the hard negative set $\mathcal{N}$ (see Figure \ref{fig:MDL}).
Formally, we construct:
{\small
\begin{align}\label{eq:construct}
\mathcal{P} = \{ (i,j)| f(x_i)^\mathrm{T}f(x_j)\geq S, A(y_i,y_j)\geq T \} \nonumber \\
\mathcal{N} = \{ (k,l)| f(x_k)^\mathrm{T}f(x_l)\geq S, A(y_k,y_l)<T \}
\end{align}
}%
where $S$ is the cosine similarity (inner product) of the $pM$-th pair after sorting all $M$ pairs in an descending order according to the feature similarity
(i.e. $S$ is a similarity threshold),
and $T$ is the similarly defined threshold value for the soft multilabel agreement.
Then we formulate the soft Multilabel-guided Discriminative embedding Learning by:
{\small
\begin{align}\label{eq:MDL}
L_{MDL} = -\log \frac{\overline{P}}{\overline{P}+\overline{N}},
\end{align}
}%
where
{\small
\begin{align}
\overline{P} = \frac{1}{|\mathcal{P}|}\Sigma_{(i,j)\in\mathcal{P}} \exp(-||f(z_i)-f(z_j)||^2_2), \nonumber \\
\overline{N} = \frac{1}{|\mathcal{N}|}\Sigma_{(k,l)\in\mathcal{N}} \exp(-||f(z_k)-f(z_l)||^2_2). \nonumber
\end{align}
}%
By minimizing $L_{MDL}$,
we are learning a discriminative feature embedding using the mined positive/hard negative pairs.
Note that the construction of $\mathcal{P}$ and $\mathcal{N}$ is
dynamic during model training,
and we construct them within every batch with the up-to-date feature
embedding during model learning
(in this case, we simply replace $M$ by $M_{batch}=N_{batch}\times (N_{batch}-1)/2$
where $N_{batch}$ is the number of unlabeled images in a mini-batch).

\subsection{Cross-view consistent soft multilabel learning}\label{sec:soft multilabel_learning}
Given the soft multilabel-guided hard negative mining,
we notice that most pairs in the RE-ID problem context are the \emph{cross-view} pairs
which consist of two person images captured by different camera views \cite{2017_ICCV_asymmetric}.
Therefore, {\Koven
the soft multilabel should be consistently good across different camera views
to be cross-view comparable.
}
From a distributional perspective,
given the reference persons and the unlabeled target dataset
$\mathcal{X}$ which is collected in a given target domain,
the distribution of the \emph{comparative characteristic} should
\emph{only} depend on the distribution of the person appearance in
the target domain and be independent of its camera views.
For example, if the target domain is a cold open-air market where
customers tend to wear dark clothes,
the soft multilabels should have higher label likelihood in the entries which are corresponding to those reference persons who also wear dark,
no matter in which target camera view.
In other words, the distribution of the soft multilabel in every camera
view should be consistent with the distribution of the target domain.
Based on the above analysis,
we introduce a Cross-view consistent soft Multilabel Learning
loss\footnote{For conciseness we omit all the averaging divisions for
  the outer summations in our losses.}:
{\small
\begin{align}
L_{CML} = \Sigma_{v} d(\mathbb{P}_v(y), \mathbb{P}(y))^2
\end{align}
}%
where $\mathbb{P}(y)$ is the soft multilabel distribution in the dataset $\mathcal{X}$,
$\mathbb{P}_v(y)$ is the soft multilabel distribution in the $v$-th camera view in $\mathcal{X}$,
and $d(\cdot, \cdot)$ is the distance between two distributions.
We could use any distributional distance, e.g. the KL divergence \cite{2014_NIPS_GAN} and the Wasserstein distance \cite{2017_ICML_WGAN}.
Since we empirically observe that the soft multilabel approximately follows a log-normal distribution,
in this work we adopt the simplified 2-Wasserstein distance \cite{2017_Arxiv_BEGAN,2018_TPAMI_WCNN} which gives a very simple form
(please refer to the supplementary material\footnote{\url{https://kovenyu.com/papers/2019_CVPR_MAR_supp.pdf}}
for the observations of the log-normal distribution and the derivation of the simplified 2-Wasserstein distance):
{\small
\begin{align}\label{eq:VML}
L_{CML} = \Sigma_{v} ||\mu_v - \mu||_2^2 + ||\sigma_v - \sigma||_2^2
\end{align}
}%
where $\mu$/$\sigma$ is the mean/std vector of the log-soft multilabels,
 $\mu_{v}$/$\sigma_{v}$ is the mean/std vector of the log-soft multilabels in the $v$-th camera view.
The form of $L_{CML}$ in Eq. (\ref{eq:VML}) is computationally cheap
and easy-to-compute within a batch. 
We note that the camera view label is naturally available in the unsupervised RE-ID setting \cite{2017_ICCV_asymmetric,2018_ECCV_HHL},
i.e. it is typically known from which camera an image is captured.

\subsection{Reference agent learning}\label{sec:aj}

A reference agent serves to represent a unique reference person
in the feature embedding like a compact ``feature summarizer''.
Therefore, the reference agents should be mutually discriminated from each other
while each of them should be representative of all the corresponding person images.
Considering that the reference agents are compared within the soft multilabel function $l(\cdot)$,
we formulate the Agent Learning loss as:
{\small
\begin{align}\label{eq:AL}
L_{AL} =\Sigma_k -\log  l(f(z_k), \{a_i\})^{(w_k)} =\Sigma_k -\log \frac{\exp(a_{w_k}^\mathrm{T}f(z_k))}{\Sigma_{j} \exp(a_j^\mathrm{T}f(z_k))}
\end{align}
}%
where $z_k$ is the $k$-th person image in the auxiliary dataset with its label $w_k$.

By minimizing $L_{AL}$,
we not only learn discriminatively the reference agents,
{\Koven but also endow the feature embedding with basic discriminative power
for the soft multilabel-guided hard negative mining.
Moreover, it reinforces implicitly the validity of the soft multilabel function $l(\cdot)$.
Specifically, in the above $L_{AL}$, the soft multilabel function
learns to assign a reference person image $f(z_k)$ with a soft multilabel $\hat{y}_k=l(f(z_k),\{a_i\}_{i=1}^{N_p})$
by comparing $f(z_k)$ to all agents,
with the learning goal that $\hat{y}_k$ should have minimal cross-entropy with (i.e. similar enough to) the ideal one-hot label $\hat{w}_k=[0, \cdots, 0, 1, 0, \cdots, 0]$ which could produce the \emph{ideal soft multilabel agreement},
i.e. $A(\hat{w}_i, \hat{w}_j)=1$ if $z_i$ and $z_j$ are the same person and $A(\hat{w}_i,\hat{w}_j)=0$ otherwise.
However, this $L_{AL}$ is minimized for the auxiliary dataset.
To further improve the validity of the soft multilabel function for the unlabeled target dataset
(i.e. the \emph{reference comparability} between $f(x)$ and $\{a_i\}$), we propose to learn a joint embedding as follows.
}

\vspace{0.1cm}
\noindent
\textbf{Joint embedding learning for reference comparability}.
A major challenge in achieving the reference comparability is the domain shift \cite{2010_TKDE_survey},
which is caused by different person appearance distributions
between the two independent domains.
To address this challenge,
we propose to mine the \emph{cross-domain} hard negative pairs
(i.e. the pair consisting of an unlabeled person $f(x)$ and an auxiliary reference person $a_i$)
to rectify the cross-domain distributional misalignment.
Intuitively, for each reference person $a_i$,
we search for the unlabeled persons $f(x)$
that are visually similar to $a_i$.
For a joint feature embedding where the discriminative distributions are well aligned,
$a_i$ and $f(x)$ should be discriminative enough to each other despite their high visual similarity.
Based on the above discussion, we propose the Reference agent-based Joint embedding learning loss\footnote{For brevity
we omit the negative auxiliary term (i.e. $w_k\neq i$) which is for a balanced learning in both domains,
as our focus is to rectify the cross-domain distribution misalignment.}:
{\small
\begin{align}\label{eq:A-Joe}
L_{RJ} = \Sigma_i \Sigma_{j\in \mathcal{M}_i} \Sigma_{k\in \mathcal{W}_i}  [m-\lVert a_i - f(x_j)\rVert^2_2]_+ + \lVert a_i - f(z_k)\rVert^2_2
\end{align}
}%
where $\mathcal{M}_i=\{j| \lVert a_i - f(x_j)\rVert^2_2<m\}$ denotes the mined data associated with the $i$-th agent $a_i$,
$m=1$ is the agent-based margin which has been theoretically justified in \cite{2017_ACMMM_normface}
with its recommaned value 1,
$[\cdot]_+$ is the hinge function,
and
$\mathcal{W}_i=\{k| w_k=i\}$.
The center-pulling term $||a_i - f(z_k)||^2_2$ reinforces the representativeness of the reference agents
to improve the validity that $a_i$ represents a reference person in the cross-domain pairs ($a_i$, $f(x_j)$).

We formulate the Reference Agent Learning by:
{\small
\begin{align}\label{eq:AJ}
L_{RAL} = L_{AL} + \beta L_{RJ}
\end{align}
}%
where $\beta$ balances the loss magnitudes.

\subsection{Model training and testing}
To summarize, the loss objective of our deep soft \textbf{m}ultil\textbf{a}bel \textbf{r}eference learning (MAR)
is formulated by:
{\small
\begin{align}\label{eq:final}
L_{MAR} = L_{MDL} + \lambda_{1} L_{CML} + \lambda_{2} L_{RAL}
\end{align}
}%
where $\lambda_1$ and $\lambda_2$ are hyperparameters to control the relative importance of the cross-view consistent soft multilabel learning
and the reference agent learning, respectively.
We train our model end to end by the Stochastic Gradient Descent (SGD).
For testing, we compute the cosine feature similarity of each probe(query)-gallery pair,
and obtain the ranking list of the probe image against the gallery images.

\section{Experiments}

\subsection{Datasets}

\begin{figure}[t]
\begin{center}
\vspace{-0.2cm}
\includegraphics[width=0.8\linewidth,height=0.2\linewidth]{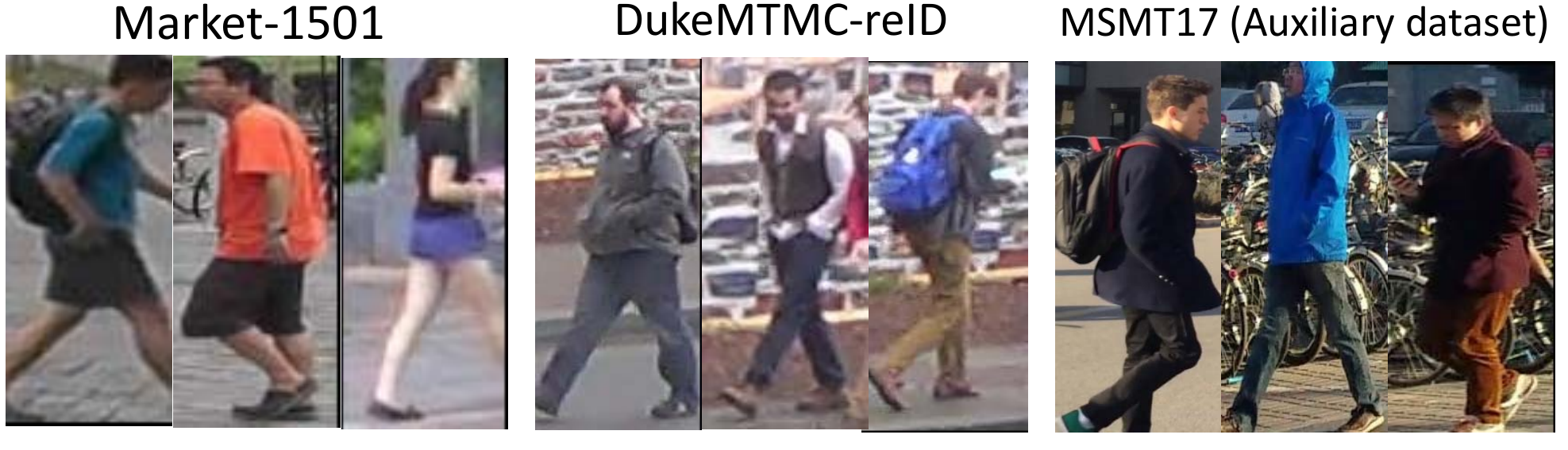}
\end{center}
\vspace{-0.5cm}
   \caption{Dataset examples.
   }
\label{fig:datasets}
\vspace{-0.68cm}
\end{figure}

\noindent
\textbf{Evaluation benchmarks}.
We evaluate our model in two widely used large RE-ID benchmarks
Market-1501 \cite{2015_ICCV_MARKET}
and DukeMTMC-reID \cite{Duke1,Duke2}.
The Market-1501 dataset has 32,668 person images of 1,501 identities.
There are in total 6 camera views.
The Duke dataset has 36,411 person images of 1,404 identities.
There are in total 8 camera views.
We show example images in Figure \ref{fig:datasets}.
We follow the standard protocol \cite{2015_ICCV_MARKET,Duke1} where the training set contains half of the identities,
and the testing set contains the other half.
We do not use any label of the target datasets during training.
The evaluation metrics are the Rank-1/Rank-5 matching accuracy and the mean average precision (MAP) \cite{2015_ICCV_MARKET}.

\vspace{0.1cm}
\noindent
\textbf{Auxiliary dataset}.
Essentially the soft multilabel represents an unlabeled person by a set of reference persons,
and therefore a high appearance diversity of the reference population would enhance the validity and capacity of the soft multilabel.
Hence, we adopt the
MSMT17 \cite{2018_CVPR_PTGAN} RE-ID dataset as the auxiliary dataset,
which has more identities (i.e. 4,101 identities) than any other RE-ID dataset and which is collected along several days
instead of a single day (different weathers could lead to different dressing styles).
There are in total 126,441 person images in the MSMT17 dataset.
Adopting the MSMT17 as auxiliary dataset enables us to evaluate
how various numbers of reference persons
(including when there are only a small number of reference persons) affect our model learning
in Sec. \ref{sec:further}.

\subsection{Implementation details}\label{sec:details}

We set batch size $B=368$, half of which randomly samples unlabeled images $x$ and the other half randomly samples $z$.
Since optimizing entropy-based loss $L_{AL}$ with the unit norm constraint has convergence issue \cite{2017_ACMMM_normface,2016_NIPS_tuplet-loss},
we follow the training method in \cite{2017_ACMMM_normface},
i.e. we first pretrain the network using only $L_{AL}$ (without enforcing the unit norm constraint)
to endow the basic discriminative power with the embedding and to determine the directions of the reference agents
in the hypersphere embedding \cite{2017_ACMMM_normface},
then we enforce the constraint to start our model learning and multiply the constrained inner products by the average inner product value in the pretraining.
We set $\lambda_1 = 0.0002$ which controls the relative importance of soft multilabel learning
and $\lambda_2 = 50$ which controls the relative importance of agent reference learning.
We show an evaluation on $\lambda_1$ and $\lambda_2$ in Sec. \ref{sec:further}.
We set the mining ratio $p$ to 5\textperthousand and set $\beta=0.2$.
Training is on four Titan X GPUs and the total time is about 10 hours.
We leave the evaluations on $p$/$\beta$ and further details in the supplementary material due to space limitation.


\subsection{Comparison to the state of the art}

\begin{table}[t]
\centering
\scriptsize
\caption{\label{tb:stoa_Market}
Comparison to the state-of-the-art unsupervised results in the Market-1501 dataset.
\red{\textbf{Red}} indicates the best and \blue{\textbf{Blue}} the second best.
Measured by \%.
}
\begin{tabular}{c|c|cc|c}
\hline
\multirow{2}*{Methods} & \multirow{2}*{Reference} & \multicolumn{3}{c}{Market-1501} \\
\cline{3-5}
&& rank-1 & rank-5 & mAP \\
\hline
\hline
LOMO \cite{2015_CVPR_LOMO}& CVPR'15 & 27.2 & 41.6  & 8.0 \\
BoW \cite{2015_ICCV_MARKET}& ICCV'15 & 35.8 & 52.4  & 14.8 \\
DIC \cite{Dic}& BMVC'15 & 50.2 & 68.8  & 22.7 \\
ISR \cite{ISR}& TPAMI'15 & 40.3 & 62.2 & 14.3 \\
UDML \cite{2016_CVPR_tDIC}& CVPR'16 & 34.5 & 52.6 & 12.4 \\
\hline
CAMEL \cite{2017_ICCV_asymmetric}& ICCV'17 & 54.5 & 73.1  & 26.3 \\
PUL \cite{2017_Arxiv_PUL}& ToMM'18 & 45.5 & 60.7  & 20.5 \\
\hline
TJ-AIDL \cite{2018_CVPR_transferable}& CVPR'18 & 58.2 & 74.8  & 26.5 \\
PTGAN \cite{2018_CVPR_PTGAN}& CVPR'18 & 38.6 & 57.3  & 15.7 \\
SPGAN \cite{2018_CVPR_SPGAN}& CVPR'18 & 51.5 & 70.1  & 27.1 \\
HHL \cite{2018_ECCV_HHL}& ECCV'18 & \blue{\textbf{62.2}} & \blue{\textbf{78.8}} & 31.4 \\
DECAMEL \cite{2019_TPAMI_DECAMEL} & TPAMI'19 &60.2 &76.0 &\blue{\textbf{32.4}}\\
\hline
MAR & This work & \red{\textbf{67.7}} & \red{\textbf{81.9}} & \red{\textbf{40.0}} \\
\hline
\end{tabular}
\vspace{-0.2cm}
\end{table}

\begin{table}[t]
\centering
\scriptsize
\caption{\label{tb:stoa_Duke}
Comparison to the state-of-the-art unsupervised results in the DukeMTMC-reID dataset. Measured by \%.
}
\begin{tabular}{c|c|cc|c}
\hline
\multirow{2}*{Methods} & \multirow{2}*{Reference} & \multicolumn{3}{c}{DukeMTMC-reID} \\
\cline{3-5}
&& rank-1 & rank-5 & mAP \\
\hline
\hline
LOMO  \cite{2015_CVPR_LOMO}& CVPR'15 & 12.3 & 21.3  & 4.8 \\
BoW \cite{2015_ICCV_MARKET}& ICCV'15 & 17.1 & 28.8  & 8.3 \\
UDML \cite{2016_CVPR_tDIC}& CVPR'16 & 18.5 & 31.4  & 7.3 \\
\hline
CAMEL \cite{2017_ICCV_asymmetric}& ICCV'17 & 40.3 & 57.6  & 19.8 \\
PUL \cite{2017_Arxiv_PUL} & ToMM'18 & 30.0 & 43.4 & 16.4 \\
\hline
TJ-AIDL \cite{2018_CVPR_transferable} & CVPR'18 & 44.3 & 59.6  & 23.0 \\
PTGAN  \cite{2018_CVPR_PTGAN}& CVPR'18 & 27.4 & 43.6  & 13.5 \\
SPGAN  \cite{2018_CVPR_SPGAN}& CVPR'18 & 41.1 & 56.6  & 22.3 \\
HHL \cite{2018_ECCV_HHL} & ECCV'18 & \blue{\textbf{46.9}} & \blue{\textbf{61.0}} & \blue{\textbf{27.2}} \\
\hline
MAR & This work &  \red{\textbf{67.1}} & \red{\textbf{79.8}} & \red{\textbf{48.0}} \\
\hline
\end{tabular}
\vspace{-0.2cm}
\end{table}

We compare our model with the state-of-the-art unsupervised RE-ID
models including: \textbf{(1)} the \emph{hand-crafted feature representation}
based models
LOMO \cite{2015_CVPR_LOMO}, BoW \cite{2015_ICCV_MARKET}, DIC
\cite{Dic}, ISR \cite{ISR} and UDML \cite{2016_CVPR_tDIC};
\textbf{(2)} the \emph{pseudo label learning} based models CAMEL \cite{2017_ICCV_asymmetric},
DECAMEL \cite{2019_TPAMI_DECAMEL} and PUL \cite{2017_Arxiv_PUL};
and \textbf{(3)} the \emph{unsupervised domain adaptation} based
models TJ-AIDL \cite{2018_CVPR_transferable}, PTGAN
\cite{2018_CVPR_PTGAN}, SPGAN \cite{2018_CVPR_SPGAN} and HHL
\cite{2018_ECCV_HHL}.
We show the results in Table \ref{tb:stoa_Market} and Table \ref{tb:stoa_Duke}.

From Table \ref{tb:stoa_Market} and Table \ref{tb:stoa_Duke}
we observe that our model could significantly outperform the state-of-the-art methods.
Specifically, our model achieves an improvement over the current state
of the art (HHL in ECCV'18)
by \textbf{20.2\%/20.8\%} \jason{on} Rank-1 accuracy/MAP in the DukeMTMC-reID dataset
and by \textbf{5.5\%/8.6\%} in the Market-1501 dataset.
This observation validates the effectiveness of MAR.

\vspace{0.1cm}
\noindent
\textbf{Comparison to the hand-crafted feature representation based models}.
The performance gaps are most significant when comparing our model to the hand-crafted feature based models
\cite{2015_CVPR_LOMO,2015_ICCV_MARKET,Dic,ISR,2016_CVPR_tDIC}.
The main reason is that these early works
are mostly based on heuristic design, and thus they could not learn optimal discriminative features.

\vspace{0.1cm}
\noindent
\textbf{Comparison to the pseudo label learning based models}.
Our model significantly outperforms the pseudo label learning based unsupervised RE-ID models \cite{2017_ICCV_asymmetric,2017_Arxiv_PUL}.
A key reason is that
our soft multilabel reference learning could exploit the auxiliary reference information to mine the potential discriminative information
that is hardly detectable when directly comparing the visual features of a pair of visually similar persons.
In contrast,
the pseudo label learning based models assign the pseudo label
by direct comparison of the visual features (e.g. via K-means clustering \cite{2017_ICCV_asymmetric,2017_Arxiv_PUL}),
rendering them blind to the potential discriminative information.

\vspace{0.1cm}
\noindent
\textbf{Comparison to the unsupervised domain adaptation based models}.
Compared to the unsupervised domain adaptation based RE-ID models \cite{2018_CVPR_PTGAN,2018_CVPR_SPGAN,2018_ECCV_HHL,2018_CVPR_transferable},
our model achieves superior performances.
A key reason is that these models only focus on
transfering/adapting the discriminative knowledge from the source domain
but ignore the discriminative label information mining in the unlabeled target domain.
The discriminative knowledge in the source domain could be less effective in the target domain even after adaptation,
because the discriminative clues can be drastically different.
In contrast, our model mines the discriminative information in the unlabeled target data,
which contributes direct effectiveness to the target RE-ID task.

\subsection{Ablation study}\label{sec:ablation}

\begin{table}[t]
\centering
\scriptsize
\caption{\label{tb:ablation}
Ablation study.
 Please refer to the text in Sec. \ref{sec:ablation}.
}
\begin{tabular}{c|ccc|c}
\hline
\multirow{2}*{Methods}  & \multicolumn{4}{c}{Market-1501} \\
\cline{2-5}
& rank-1 & rank-5 & rank-10 & mAP \\
\hline
Pretrained (source only) &46.2 & 64.4 & 71.3 & 24.6 \\
\hline
Baseline (feature-guided) & 44.4 & 62.5 & 69.8 & 21.5 \\
\hline
MAR w/o $L_{CML}$ &60.0&75.9&81.9&34.6 \\
MAR w/o $L_{CML}$\&$L_{RAL}$ &53.9&71.5&77.7&28.2 \\
MAR w/o $L_{RAL}$ &59.2&76.4&82.3&30.8\\
\hline
MAR & \textbf{67.7} & \textbf{81.9}& \textbf{87.3} &\textbf{40.0}\\
\hline
\hline
\multirow{2}*{Methods}  & \multicolumn{4}{c}{DukeMTMC-reID} \\
\cline{2-5}
 & rank-1 & rank-5 & rank-10  & mAP \\
 \hline
Pretrained (source only) & 43.1 & 59.2 & 65.7 & 28.8 \\
\hline
Baseline (feature-guided) & 50.0 & 66.4 & 71.7 & 31.7\\
\hline
MAR w/o $L_{CML}$ &63.2 & 77.2 & 82.5 & 44.9\\
MAR w/o $L_{CML}$\&$L_{RAL}$ & 60.1 & 73.0 & 78.4 & 40.4\\
MAR w/o $L_{RAL}$ &57.9 & 72.6 & 77.8 & 37.1\\
\hline
MAR & \textbf{67.1} & \textbf{79.8} & \textbf{84.2} & \textbf{48.0}\\
\hline
\end{tabular}
\vspace{-0.2cm}
\end{table}

We perform an ablation study to demonstrate
(1) the effectiveness of the soft multilabel guidance
and
(2) the indispensability of the
cross-view consistent soft multilabel learning and the reference agent learning to MAR.
For (1), we adopt the \emph{pretrained} model {\Koven (i.e. only trained by $L_{AL}$ using the auxiliary source MSMT17 dataset to have basic discriminative power, as mentioned in Sec. \ref{sec:details})}.
We also adopt a \emph{baseline} model that is feature similarity-guided instead of soft multilabel-guided.
Specifically, after the same pretraining procedure,
we replace the soft multilabel agreement with the feature similarity,
i.e. in the hard negative mining we partition the mined similar pairs into two halves by a threshold of feature similarity rather than soft multilabel agreement,
and thus regard the high/low similarity half as positive set $\mathcal{P}$/hard negative set $\mathcal{N}$.
For (2), we remove the $L_{CML}$ or $L_{RAL}$.
We show the results in Table \ref{tb:ablation}.

\vspace{0.1cm}
\noindent
\textbf{Effectiveness of the soft multilabel-guided hard negative mining}.
Comparing MAR to the pretrained model where the soft multilabel-guided hard negative mining is missing,
we observe that MAR significantly improves the pretrained model
(e.g. \jason{on} Market-1501/DukeMTMC-reID, MAR improves the pretrained model by 21.5\%/24.0\% \jason{on} Rank-1 accuracy).
{\Koven This is because the pretrained model is only discriminatively trained on the auxiliary source dataset without mining the discriminative information in the unlabeled target dataset, so that it is only discriminative on the source dataset but not the target.}
This comparison demonstrates the effectiveness of the soft multilabel-guided hard negative mining.

\vspace{0.1cm}
\noindent
\textbf{Effectiveness of the soft multilabel agreement guidance}.
Comparing MAR to the baseline model,
we observe that MAR also significantly outperforms the similarity-guided hard negative mining baseline model.
(e.g. \jason{on} Market-1501/DukeMTMC-reID, MAR outperforms the similarity-guided hard negative mining baseline by 23.3\%/17.1\% \jason{on} Rank-1 accuracy).
Furthermore, even when the soft multilabel learning and reference agent learning losses are missing
(i.e. ``MAR w/o $L_{CML}\&L_{RAL}$'' where the soft multilabel is much worse than MAR),
the soft multilabel-guided model still outperforms the similarity-guided model by 14.8\%/7.9\% \jason{on} Rank-1 accuracy \jason{on} Market-1501/DukeMTMC.
These demonstrate the effectiveness of the soft multilabel guidance.

\vspace{0.1cm}
\noindent
\textbf{Indispensability of the soft multilabel learning and the reference agent learning}.
When the cross-view consistent soft multilabel learning loss is absent,
the performances drastically drop
(e.g. drop by 7.7\%/5.4\% \jason{on R}ank-1 accuracy/MAP in the Market-1501 dataset).
This is mainly because optimizing $L_{CML}$ improves the soft multilabel comparability of the cross-view pairs \cite{2017_ICCV_asymmetric},
giving more accurate judgement in the positive/hard negative pairs.
Hence, the cross-view consistent soft multilabel learning is indispensable in MAR.
When the reference agent learning loss is also absent,
the performances further drop (e.g. drop by 13.8\%/11.8\% \jason{on R}ank-1/MAP in the Market-1501 dataset).
This is because in the absence of the reference agent learning,
the soft multilabel is {\Koven learned via comparing to the less valid reference agents (only pretrained).}
This observation validates the importance of the reference agent learning.
\subsection{Visual results and insight}\label{sec:qualitative}

\begin{figure}[t]
\begin{center}
\includegraphics[width=0.99\linewidth]{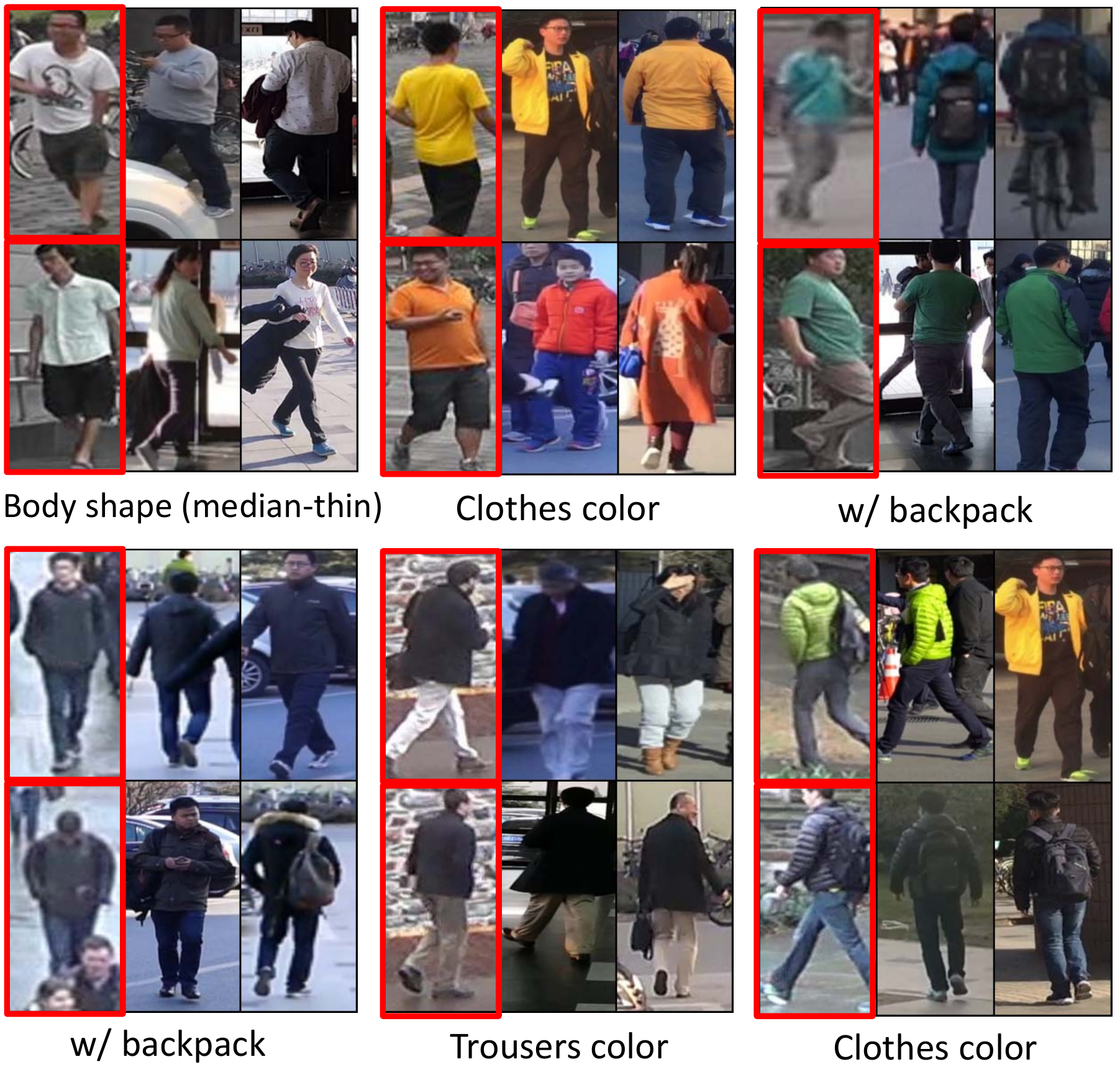}
\vspace{-0.1cm}
\end{center}
   \caption{Visual results of the soft multilabel-guided hard negative mining.
   Each pair surrounded by the \red{red box} is the similar pair mined by our model with the lowest soft multilabel agreements,
   and the images on their right are the reference persons corresponding to the first/second maximal soft multilabel entries.
   The first row is from the Market-1501 and the second from DukeMTMC-reID.
   We highlight the discovered fine-grained discriminative clues in the bottom text for each pair.
   Please view in the screen and zoom in.
   }
\label{fig:qualitative}
\end{figure}

To demonstrate how the proposed soft multilabel reference learning works,
in Figure \ref{fig:qualitative} we show the similar target pairs with the lowest soft multilabel agreements
(i.e. the mined soft multilabel-guided hard negative pairs) mined by our trained model.
We make the following observations:

(1) For an unlabeled person image $x$, the maximal entries (label likelihood) of the learned soft multilabel
are corresponding to the reference persons that are highly visually similar to $x$,
i.e. the soft multilabel represents an unlabeled person mainly by visually similar reference persons.

(2) For a pair of visually similar but unlabeled person images,
the soft multilabel reference learning works by discovering potential fine-grained discriminative clues.
For example, in the upper-right pair in Figure \ref{fig:qualitative},
the two men are dressed similarly.
A potential fine-grained discriminative clue is whether they have a backpack.
For the man taking a backpack, the soft multilabel reference learning assigns maximal
label likelihood to two reference persons who also take backpacks,
while for the other man the two reference persons do not take backpacks, either.
As a result, the soft multilabel agreement is very low, giving a judgement that this is a hard negative pair.
We highlight the discovered fine-grained discriminative clues in the bottom of every pair.

These observations lead us to conclude that the soft multilabel reference learning
distinguishes visually similar persons by giving high label likelihood to different reference persons
to produce a low soft multilabel agreement.

\subsection{Further evaluations}\label{sec:further}

\begin{figure}[t]
\begin{center}
\subfigure[Market-1501]{
\includegraphics[width=0.4\linewidth]{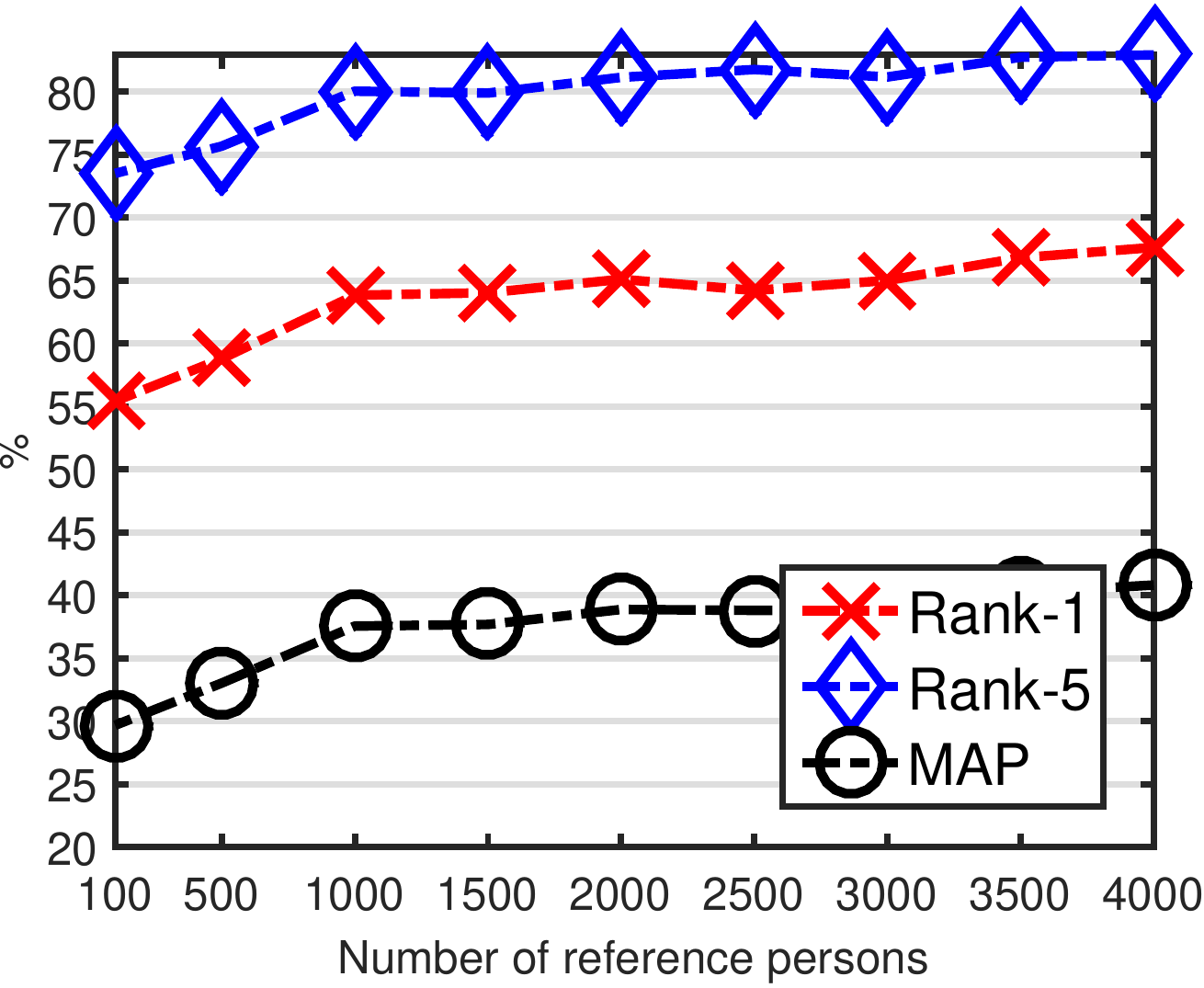}\label{fig:n_ref_vary_market}
}
\vspace{-0.4cm}
\subfigure[DukeMTMC-reID]{
\includegraphics[width=0.4\linewidth]{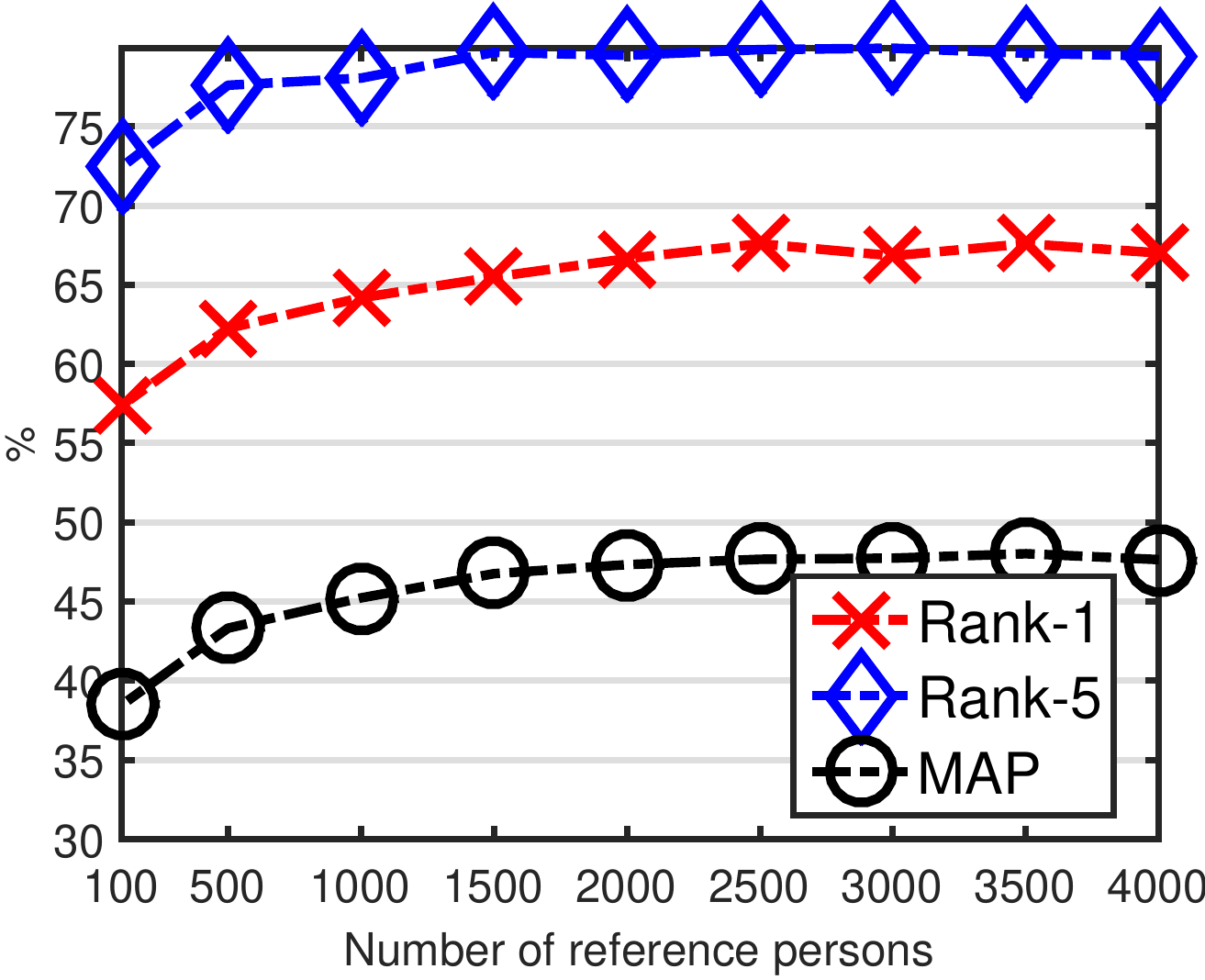}\label{fig:n_ref_vary_duke}
}
\end{center}
   \caption{Evaluation on different numbers of reference persons.
   }
\label{fig:n_ref_vary}
\vspace{-0.4cm}
\end{figure}

\noindent
\textbf{Various numbers of reference persons}.
We evaluate how the number of reference persons affect our model learning.
In particular, we vary the number by using only the first $N_u$ reference persons
(except that we keep all data used in $L_{AL}$ to guarantee that the basic discriminative power is not changed).
We show the results in Figure \ref{fig:n_ref_vary}.

{\Koven
From Figure \ref{fig:n_ref_vary_market} we observe that:
(1) Empirically, the performances become stable when the number of reference persons are larger than 1,500, which is approximately \emph{two times of} the number of the training persons in both datasets (750/700 training persons in Market-1501/DukeMTMC-reID).
This indicates that MAR does not necessarily require a very large reference population but a median size, e.g. two times of the training persons.
(2) When there are only a few reference persons (e.g. 100),
the performances drop drastically due to the poor soft multilabel representation capacity of the small reference population.
In other words, this indicates that MAR could not be well learned using a very small auxiliary dataset.
}

\begin{figure}[t]
\begin{center}
\subfigure[$\lambda_1$]{
\includegraphics[width=0.4\linewidth]{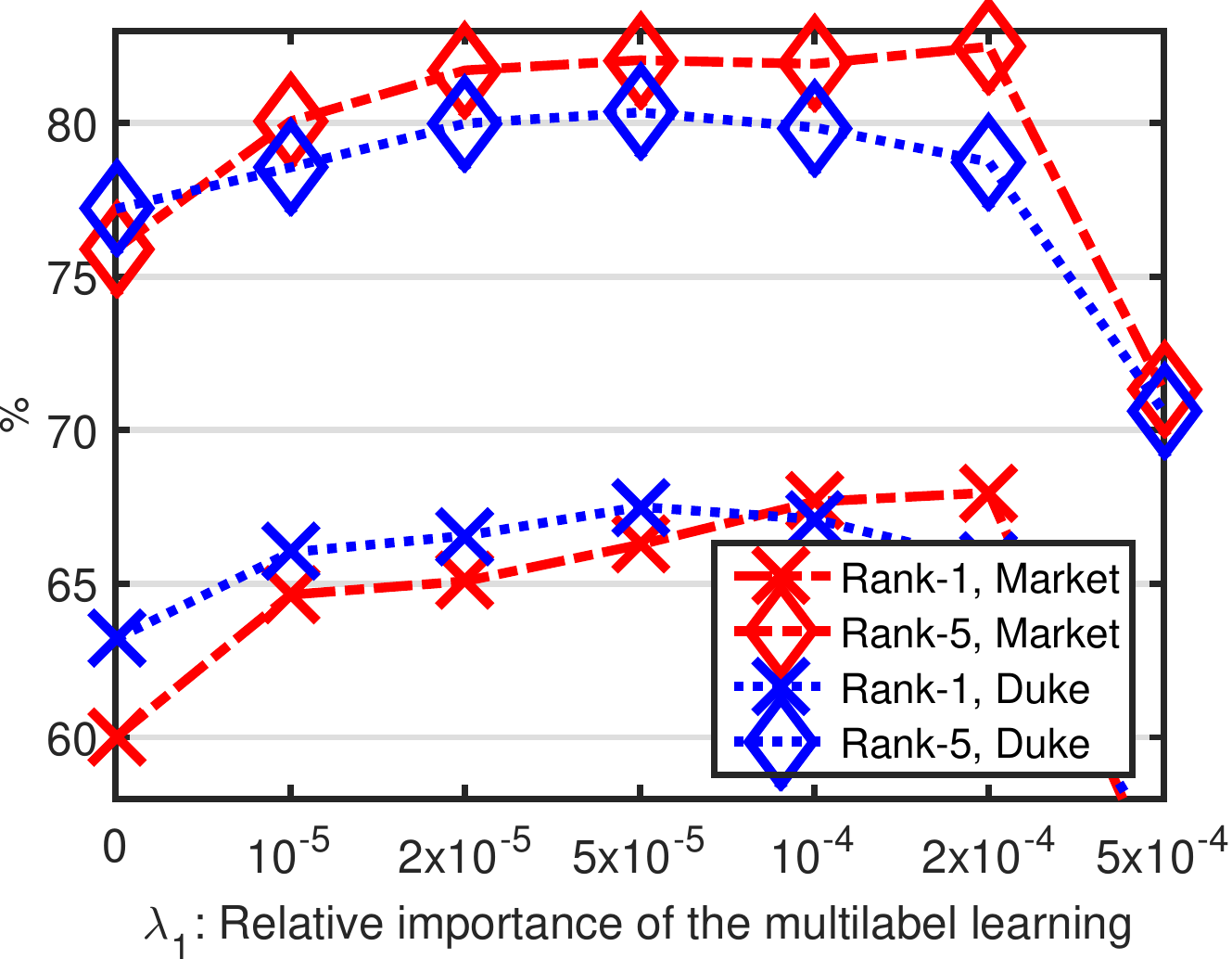}\label{fig:lambda1_vary}
}
\vspace{-0.4cm}
\subfigure[$\lambda_2$]{
\includegraphics[width=0.4\linewidth]{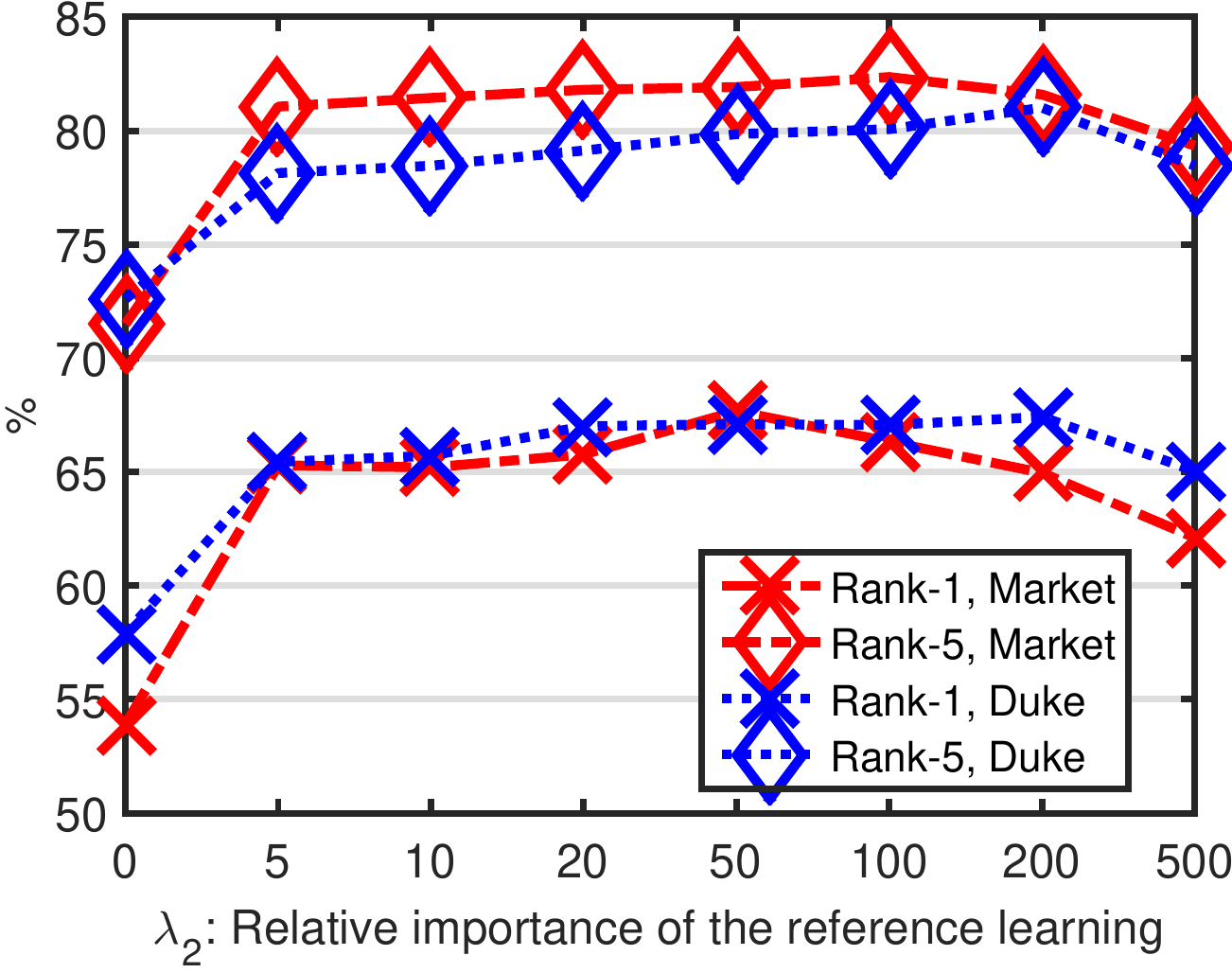}\label{fig:lambda2_vary}
}
\end{center}
   \caption{
   Evaluation on important hyperparameters.
   For (a) we fix $\lambda_2=50$ and for (b) we fix $\lambda_1=0.0002$.
   }
\label{fig:lambda_vary}
\vspace{-0.4cm}
\end{figure}

\vspace{0.1cm}
\noindent
\textbf{Hyperparameter evaluations}.
We evaluate how $\lambda_1$ (which controls the relative importance of the soft multilabel learning)
and $\lambda_2$ (the relative importance of the reference agent learning) affect our model learning.
We show the results in Figure \ref{fig:lambda_vary}.
From Figure \ref{fig:lambda_vary} we observe that
our model learning is stable within a wide range for both hyperparameters
(e.g. $2\times 10^{-5}<\lambda_1<5\times 10^{-4}$ and $10<\lambda_2<200$),
although both of them should not be too large to overemphasize the soft multilabel/reference agent learning.

%
%

\section{Conclusion}

In this work we demonstrate the effectiveness of utilizing auxiliary source RE-ID data
for mining the potential label information latent in the unlabeled target RE-ID data.
Specifically, we propose MAR which enables simultaneously
the soft multilabel-guided hard negative mining, the cross-view consistent soft multilabel learning and the reference agent learning
in a unified model.
In MAR, we leverage the soft multilabel
for mining the latent discriminative information
that cannot be discovered by direct comparison of the \emph{absolute} visual features in the unlabeled RE-ID data.
To enable the soft multilabel-guided hard negative mining in MAR,
we simultaneously optimize the cross-view consistent soft multilabel learning and the reference agent learning.
Experimental results in two benchmarks validate the effectiveness of the proposed MAR
and each learning component of MAR.

{\small
\bibliographystyle{ieee}
\bibliography{Koven}
}

\end{document}